# E2E-MLT - an Unconstrained End-to-End Method for Multi-Language Scene Text

Michal Bušta[1], Yash Patel[2], Jiri Matas[1]

[1]Center for Machine Perception, Department of Cybernetics
Czech Technical University, Prague, Czech Republic
[2] Robotics Institute, Carnegie Mellon University, Pittsburgh, USA
`bustam@fel.cvut.cz, yashp@andrew.cmu.edu, matas@cmp.felk.cvut.cz`

**Abstract.** An end-to-end trainable (fully differentiable) method for multi-language scene text localization and recognition is proposed. The approach is based on a single fully convolutional network (FCN) with shared layers for both tasks.

E2E-MLT is the first published multi-language OCR for scene text. While trained in multi-language setup, E2E-MLT demonstrates competitive performance when compared to other methods trained for English scene text alone. The experiments show that obtaining accurate multi-language multi-script annotations is a challenging problem.

## 1 Introduction

Scene text localization and recognition, a.k.a. photo OCR or text detection and recognition in the wild, is a challenging open computer vision problem. Applications of photo OCR are diverse, from helping the visually impaired to data mining of street-view-like images for information used in map services and geographic information systems. Scene text recognition finds its use as a component in larger integrated systems such as those for autonomous driving, indoor navigations and visual search engines.

The growing cosmopolitan culture in modern cities often generates environments where multi-language text co-appears in the same scene (Fig 1), triggering a demand for a unified multi-language scene text system. The need is also evident from the high interest in the ICDAR competition on multi-language text [33].

Recent advances in deep learning methods have helped in improving both the text localization [14, 26, 31, 43] and text recognition [5, 18, 39] methods significantly. However, from the point of view of multi-language capabilities, these methods fall short in following aspects: (1) the evaluation is limited to English text only and the methods are not trained in a multi-language setup, (2) they solve text localization and recognition as two independent problems [14, 16] or make use of multiple networks to solve individual problems [5, 30] and (3) the existing OCR methods are lacking the ability to handle rotated or vertical text instances.

Multi-language scene text poses specific challenges. Firstly, the data currently publicly available for non-English scene text recognition is insufficient for training deep architectures. Individual languages have specific challenges, for example, CHINESE and JAPANESE have a high number of characters and there is a substantial number of instances where the text is vertical. BANGLA scene text is mostly hand written.

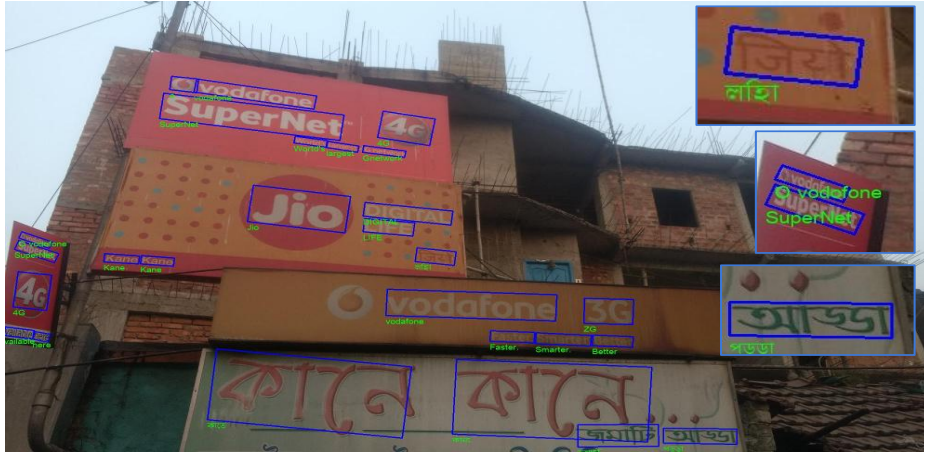

Fig. 1: Text in multiple languages appearing in a scene. The proposed E2E-MLT method localizes words, predicts the scripts and generates a text transcription for each bounding box.

In this paper, we introduce E2E-MLT, an end-to-end trainable FCN based method with shared convolutional layers for multi-language scene text. E2E-MLT handles multi-language scene text localization, text recognition and script identification using a single fully convolutional network. The method has been trained for the following languages: ARABIC, BANGLA, CHINESE, JAPANESE, KOREAN, LATIN and is capable of recognizing 7,500 characters (compared to less than 100 in English [5,16,39]) and does not make use of any fixed dictionary of words [16,36]. With E2E-MLT, we make the following contributions:

- We experimentally demonstrate that script identification is not required to recognize multi-language text. Unlike competing methods [12,33], E2E-MLT performs script identification from the OCR output using a simple majority voting mechanism over the predicted characters.

- E2E-MLT is capable of recognizing highly rotated and vertical text instances, which is achieved by using a $cos(x) - sin(x)$ representation of the angle.

- We validate that FCN based architecture is capable of handling multi-language text detection and recognition. E2E-MLT is the first published multi-language OCR which works well across six languages.

- We provide the statistics of co-occurring languages at the image level and word level on the ICDAR RRC-MLT 2017 [33] dataset. These statistics demonstrate that characters from different languages can not only co-occur in same image, but also co-occur within the same word.

- We publicly release a large scale synthetically generated dataset for training multi-language scene text detection, recognition and script identification methods.

The rest of the paper is structured as follows. In Section 2, previous work is reviewed. The method is described in Section 3 and experimentally evaluated in Section 4. Finally, conclusions are drawn is Section 5.

## 2  Related Work

### 2.1  Scene Text Localization

Scene text localization is the first step in standard text-spotting pipelines. Given a natural scene image, the objective is to obtain precise word level bounding boxes or segmentation maps.

Conventional methods such as [8, 34, 35] rely on manually designed features to capture the properties of scene text. Generally these methods seek character candidates via extremal region extraction or edge detection. Character-centric deep learning based method [18] makes use of a CNN [23] for image patches (obtained by sliding window) to predict text/no-text score, a character and a bi-gram class.

Jaderberg *et al.* [16] proposed a multi-stage word-centric method where horizontal bounding box proposals are obtained by aggregating the output of Edge Boxes [45] and Aggregate Channel Features [7]. The proposals are filtered using a Random Forest [4] classifier. As post-processing a CNN regressor is used to obtain fine-grained bounding boxes. Gupta *et al.* [14] drew inspiration from YOLO object detector [37] and proposed a fully-convolutional regression network trained on synthetic data for performing detection and regression at multiple scales in an image.

Tian *et al.* [43] use a CNN-RNN joint model to predict the text/no-text score, the y-axis coordinates and the anchor side-refinement. A similar approach [26] adapts the SSD object detector [29] to detect horizontal bounding boxes. Ma *et al.* [31] detects text of different orientations by adapting the Faster-RCNN [10] architecture and adding 6 hand-crafted rotations and 3 aspects.

A two-staged method for word or line level localization is proposed by Zhou *et al.* [44]. Following the architecture design principle of *U-Shape* [38] (a fully convolutional network with gradually merged features from different layers) is used. The text proposals obtained are then processed using NMS for final output.

As mentioned earlier, all of these methods deal with English text only. Methods trained for multi-language setup are described in ICDAR RRC-MLT 2017 [33]. *SCUT-DLVClab* trains two models separately, the first model predicts bounding box detections and second model classifies the detected bounding box into one of the script classes or background. *TH-DL* use a modified FCN with residual connections for generating text-proposals and a Fast-RCNN [10] for detection. GoogleLeNet architecture [42] is used for script-identification.

## 2.2   Scene Text Recognition

The objective of text recognition methods is to take the cropped word image and generate the transcription of the word present. Scene text recognition has been widely studied for English text. Jaderberg *et al.* [16] train a VGG-16 [41] based CNN on 9 million synthetic images to classify a cropped word image as one of the words in a dictionary. The dictionary contains roughly 90 000 English words and the words of the training and test set. Any word outside the dictionary is ignored.

Shi *et al.* [39] generates one sequence of characters per image by training a fully-convolutional network with a bidirectional LSTM using the Connectionist Temporal Classification (CTC) [13]. Unlike the OCR of proposed E2E-MLT, both [16,39] resize the source cropped word image to the fixed-sized matrix of $100 \times 32$ pixels regardless of the number of characters present or original aspect ratio.

Lee *et al.* [24] present a recursive recurrent network with soft-attention mechanism for lexicon-free scene text recognition. This method makes use of recursive CNNs with weight sharing to reduce the number of parameters. A RNN is used on top of convolutional features to automatically learn character level language model.

The aforementioned methods only deal with English text, where the number of characters is limited. Methods like [16,18] approach the problem from close dictionary perspective where any word outside the dictionary is ignored. Such setting is not feasible in multi-language scenario where the number of characters and possible set of words are very high. Text recognition in E2E-MLT is open-dictionary (in terms of set of words) and does not require any language specific information.

## 2.3   End-to-End Scene Text Methods

Recent end-to-end methods address both detection and recognition in a single pipeline. Given an image, the objective is to predict precise word level bounding boxes and corresponding text transcriptions.

Li *et al.* [25] make use of convolutional recurrent network to concurrently generate text bounding boxes and corresponding transcriptions. An approach similar to [31] was presented by Busta *et al.* [5], where the rotation is a continuous parameter and optimal anchor box dimensions are found using clustering on training set. The image crops obtained from predicted bounding boxes are transcribed using another fully convolutional network.

Liu *et al.* [30] introduce ROIRotate to extract the high-level features from shared convolutional layers and feed it to two separate branches for performing localization and recognition. The text localization branch utilizes the features from lower level and higher level layers to incorporate small text regions. The text recognition branch consists of multiple convolutional layers followed by bi-directional LSTM. The base network and the two branches are optimized jointly.

The idea of our approach for scene text detection and recognition is inspired by that of Busta *et al.* [5]: that is take the state of the art object detector, modify it to handle specifics of scene text and address both localization and recognition using a single pipeline. Unlike [5], E2E-MLT does not make use of multiple networks, is end-to-end trainable, shares convolutional features for both tasks and is trained in a multi-language setup for six languages.

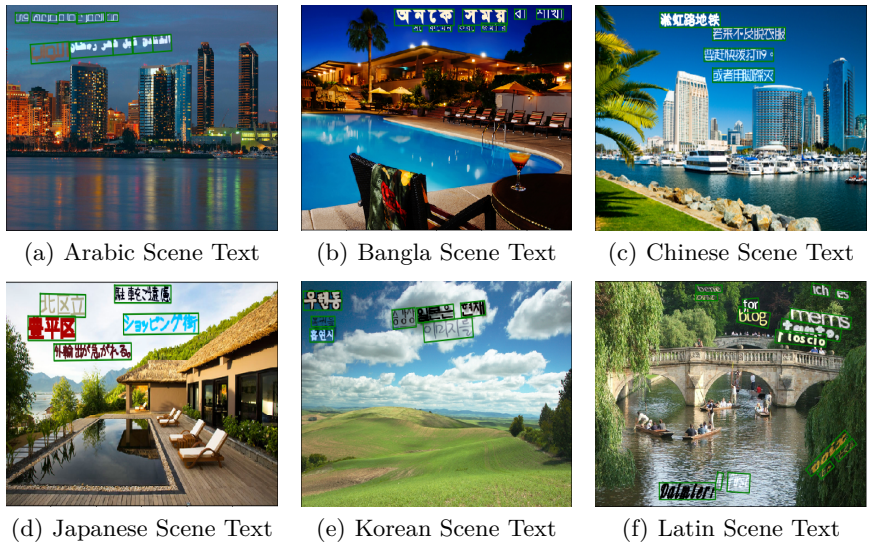

| (a) Arabic Scene Text | (b) Bangla Scene Text | (c) Chinese Scene Text |
| (d) Japanese Scene Text | (e) Korean Scene Text | (f) Latin Scene Text |

Fig. 2: Images from the Synthetic Multi-Language in Natural Scenes Dataset.

## 3    Method

Given a scene image containing instances of multi-language text, E2E-MLT obtains text localizations, generates text transcription and script class for each detected region. The structure of E2E-MLT is shown in Fig. 3. The model is optimized in an end-to-end fashion.

### 3.1    Multi-Language Synthetic Data Generation

As mentioned earlier, existing multi-language scene text datasets do not provide sufficient data for deep network training. The largest such dataset is presented in ICDAR RRC-MLT 2017 [33]. This dataset consists of 7,200 training and 1,800 validation images for six languages. To overcome the problem of limited training data, we generate a synthetic multi-language text in natural scene dataset.

We adapt the framework proposed by Gupta *et al.* [14] to a multi-language setup. The framework generates realistic images by overlaying synthetic text over existing natural background images and it accounts for 3D scene geometry. Gupta *et al.* [14] proposed the following approach for scene-text image synthesis:

- Text in the real world usually appears in well-defined regions, which can be characterized by uniform color and texture. This is achieved by thresholding gPb-UCM contour hierarchies [2] using efficient graph-cut implementation [3]. This provides prospective segmented regions for rendering text.

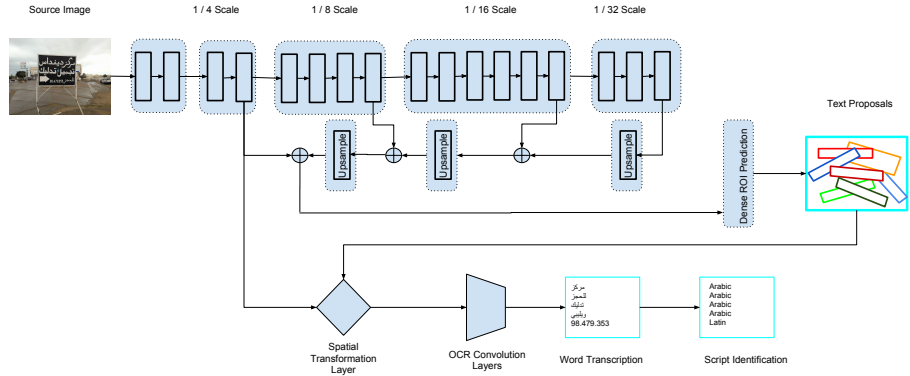

Fig. 3: E2E-MLT overview: Dense text proposals are generated and filtered by the text localization branch of E2E-MLT. Preserving the aspect ratio at bounding box level, features from shared layers are extracted. The OCR branch of E2E-MLT takes these pooled features and generates a character sequence or discards the bounding box as no-text.

- Dense estimate of depth maps of segmented regions are then obtained using [28] and then planar facets are fitted to maps using RANSAC [9]. This way normals to prospective regions for text rendering is estimated.
- Finally, the text is aligned to prospective image region for rendering. This is achieved by warping the image region to frontal-parallel view using the estimated region normals, then a rectangle is fitted to this region, and finally, the text is aligned to the larger side of this rectangle.

The generated dataset contains the same set of language classes as ICDAR RRC-MLT 2017 [33]: Arabic, Bangla, Chinese, Japanese, Korean, Latin. The **Synthetic Multi-Language in Natural Scene Dataset** contains text rendered over natural scene images selected from the set of 8,000 background images collected by [14]. Annotations include word level and character level text bounding boxes along with the corresponding transcription and language/script class. The dataset has 245,000 images with thousands of images for each language. Sample examples are shown in Fig. 2.

Note: The pipeline presented in [14] renders text in a character by character manner, which breaks down the ligature of Arabic and Bangla words. We made appropriate changes in the pipeline to handle these language specific cases. Further, in order to incorporate real world samples of highly-rotated and vertical text instances, we preserve the reading direction at the word level.

### 3.2   Overall Architecture

An overview of the E2E-MLT model is shown in Fig. 3. We make use of the FPN object detector [27] with ResNet-34 [15] as the backbone of E2E-MLT. Text in natural scene images is often relatively small compared to the entire image size so we replace the first layer of ResNet-34 with a set of 3x3 convolutions with stride 2. Similar to [27], the proposed detector works on 1/4 scale of the original image due its high memory consumption.

The architecture along with dimensionality of activation maps after different layers is detailed in Table 1. $\overline{W}$ and $\overline{H}$ are the width and height respectively of the input image. ResN-B represents ResNet block [15], IN represents Instance-Normalization [6]. $|\hat{\mathcal{A}}|$ are the number of characters that can be recognized (the union of characters in all languages, 7500 is our setup). The initial convolutional layers are shared between the both localization and recognition tasks.

| Type | Channels | Size/Stride | Dim |
|---|---|---|---|
| Con2D, IN, CReLU | 16 | $3 \times 3$ | $\overline{W} \times \overline{H}$ |
| Con2D, IN, CReLU | 32 | $3 \times 3/2$ | $\overline{W}/2 \times \overline{H}/2$ |
| Con2D, IN, ReLU | 64 | $3 \times 3$ | |
| Con2D, IN, ReLU | 64 | $3 \times 3/2$ | $\overline{W}/4 \times \overline{H}/4$ |

| OCR branch | | | | Localization branch | | | |
|---|---|---|---|---|---|---|---|
| Type | Chn. | Size/Stride | Dim | Type | Chn. | Size/Stride | Dim |
| Con2D, IN, ReLU | 128 | $3 \times 3$ | $W/4 \times 10$ | ResN-B x3 | 64 | $3 \times 3$ | $\frac{\overline{W}}{8} \times \frac{\overline{H}}{8}$ |
| Con2D, IN, ReLU | 128 | $3 \times 3$ | | ResN-B x4 | 128 | $3 \times 3/2$ | $\frac{\overline{W}}{16} \times \frac{\overline{H}}{16}$ |
| maxpool | | $2 \times 2/2 \times 1$ | $W/4 \times 5$ | ResN-B x6 | 256 | $3 \times 3/2$ | $\frac{\overline{W}}{32} \times \frac{\overline{H}}{32}$ |
| Con2D, IN, ReLU | 256 | $3 \times 3$ | | ResN-B x4 | 512 | $3 \times 3/2$ | $\frac{\overline{W}}{16} \times \frac{\overline{H}}{16}$ |
| Con2D, IN, ReLU | 256 | | | FPN lat. con. | 256 | $1 \times 1$ | $\frac{\overline{W}}{16} \times \frac{\overline{H}}{16}$ |
| maxpool | | $2 \times 2/2 \times 1$ | $W/4 \times 2$ | FPN lat. con. | 256 | $1 \times 1$ | $\frac{\overline{W}}{8} \times \frac{\overline{H}}{8}$ |
| Con2D, IN, ReLU×2 | 256 | $3 \times 3$ | | FPN lat. con. | 256 | $1 \times 1$ | $\frac{\overline{W}}{4} \times \frac{\overline{H}}{4}$ |
| Dropout (0.2) | | | | Dropout (0.2) | | | |
| Con2D | $|\hat{\mathcal{A}}|$ | $1 \times 1$ | $W/4 \times 1$ | Con2D | 7 | $1 \times 1$ | |
| log softmax | | | | | | | |

Table 1: FCN based E2E-MLT architecture for multi-language scene text localization and recognition.

## 3.3 Text Localization

The natural scene image is fed into the E2E-MLT architecture and multiple levels of per-pixel text score and geometries are generated. The prediction output consists of seven channels: per-feature text/no-text confidence score $r_p$, distances to the top, left, bottom and right sides of the bounding box that contains this pixel and the orientation angle parameters as $r_\theta$. To overcome the discontinuity in angle prediction, we represent it by $\sin(r_\theta)$ and $\cos(r_\theta)$. A shrunken version of the ground truth bounding box is considered as positive text region.

The joint loss function used for both localization and recognition in E2E-MLT is a linear combination of multiple loss functions: IoU loss proposed in [44], dice loss proposed in [32], CTC loss function proposed in [13] and mean squared error based loss for learning rotations (represented by $L_{angle}$). The overall loss function for end-to-end training is given by:

$$L_{final} = L_{geo} + \lambda_1 L_{angle} + \lambda_2 L_{dice} + \lambda_3 L_{CTC} \qquad (1)$$

Where $L_{geo}$ is the IoU loss function and is invariant against text-regions of different scales. The CTC loss term denoted by $L_{CTC}$ is for word level text recognition (details in Sec 3.4). $L_{angle}$ is a sum over mean squared loss obtained over $\sin(r_\theta)$ and $\cos(r_\theta)$ representations of rotation angle. We emperically observe that making use of two independent channels to predict the sin and cos of bounding box rotation angle helps with the localization of highly rotated and vertical text instances. Examples of generated bounding box outputs is shown in Tab 2.

Since the text regions are relatively small compared to the entire image size, there is a high class imbalance between foreground (text-region) and background class. The dice loss term helps the network to overcome this class imbalance. The dice loss as defined in [32]:

$$L_{dice} = \frac{2 \sum_i^N r_{p_i} g_i}{\sum_i^N r_{p_i}^2 + \sum_i^N q_i^2} \qquad (2)$$

Where the summation in Eq. 2 runs over all the feature points in last convolutional layer. $g_i \in \{0, 1\}$ is the ground truth label for a given feature point with index $i$ and $p_i \in (0, 1)$ is the corresponding from the network. For all our experimental setup, we keep $\lambda_1 = \lambda_2 = \lambda_3 = 1$ in Eq. 1.

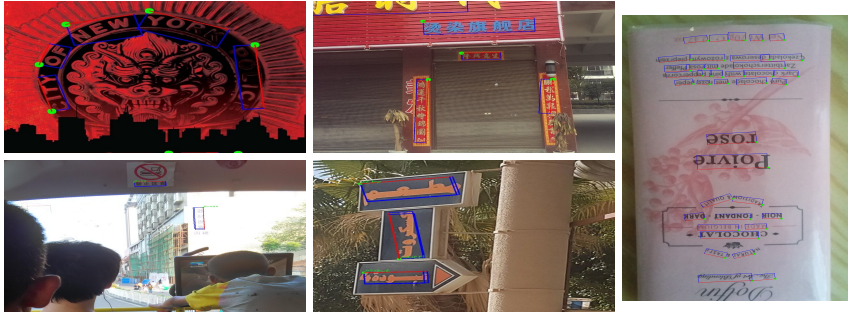

Table 2: Example of learned proposal representation (actual output of the learned network): the red line is the top line, the green circle is start of reading direction - best viewed in zoom and color

The dense proposals from the network are then filtered based on the text confidence score $r_p$ with the threshold value 0.9. Only the proposals with $r_p > 0.9$ are considered for further processing. Finally, locality aware NMS [44] is performed on filtered proposals to generate the final bounding box predictions.

### 3.4  Text Recognition

OCR branch of E2E-MLT is a FCN module for multi-language text recognition. The word level bounding boxes are predicted (as explained in Section 3.3) are compared against the ground truth bounding boxes. The predictions with the IoU of higher than 0.9 are used for training of the OCR branch of E2E-MLT.

These selected predictions are then used to estimate the warp parameters of the spatial transformer layer [17]. Note that the spatial transformer operation is applied on input image and not on the feature representation. Text recognition is highly rotation dependent. We handle rotation in separate step and make use of the spatial transformer to normalize image to scale and rotation, and thus making the learning task easier. Similar to [20], we make use of the bi-linear interpolation to compute the output values as it avoids misalignment and makes the input to the OCR module fixed height and variable width, while preserving the aspect ratio at word level.

The fully convolutional OCR module takes a variable-width feature tensor $\overline{W} \times 40 \times C$ as an input and outputs a matrix $\frac{\overline{W}}{4} \times |\hat{\mathcal{A}}|$, where $\mathcal{A}$ is the alphabet – the union of characters of all languages ( = 7500 log-Softmax outputs), and $\overline{W} = \frac{wH'}{h}$ (where $w, h$ are the width and height respectively of the text region, we fix $H' = 40$). The full network definition is provided in Tab. 1. The loss from text recognition in E2E-MLT architecture is computed using the Connectionist Temporal Classification (CTC) [13].

During all the experiments we use greedy decoding of network output. An alternative could be the use of, task and language specific techniques such as prefix decoding or decoding with language models (for the word spotting task). However, our OCR is generalized, open-dictionary and language independent.

### 3.5  Training Details

Both the text localization and OCR branches of E2E-MLT are trained together in an end-to-end fashion. A union of the ICDAR RRC-MLT 2017 [33] train dataset, the ICDAR RCTW 2017 [40] train dataset, the ICDAR 2015 [21] train dataset and Synthetic Multi-Language in Natural Scene Dataset (explained in Section. 3.1) is used for training.

We make use of Adam [22] (base learning rate = 0.0001, $\beta_1 = 0.9$, $\beta_2 = 0.999$, weight decay = 0) for optimizing over joint loss for text localization and recognition (Eq. 1) and train until the validation error converges.

## 4  Experiments

Throughout our experimental analysis, we evaluate a single model trained in multi-language setup as explained in Section 3. We do not fine-tune the model for specific datasets or task while reporting the results. Unlike [30], we evaluate the performance using a single-scale. Further, the text recognition results are reported in an unconstrained setup, that is, without making the use of any predefined lexicon (set of words).

### 4.1 Multi-Language Scene Text Recognition

First we run the analysis of the scripts co-occurrence in individual images (Tab. 3) and the scripts co-occurrence in words (Tab. 4). The script of the character is defined by the Unicode table [1]. Each character has its unique name (for example character 'A' has unicode name 'Latin Capital Letter A' therefore its script is Latin). The scripts which occur in the ICDAR MLT 2017 dataset [33] are LATIN (LAT), ARABIC (ARA), BENGALI (BENG), HANGUL (HANG), CJK, HIRAGANA (HIR), KATAKANA (KAT) and DIGIT (DIG). The rest of characters are considered to be SYMBOLS (SYM). The abbreviation CKH marks the group of CJK, HIRAGANA AND KATAKANA scripts. Tab. 4 shows that the script co-occurrence is non-trivial even on the word level. The OCR module in a practical application should satisfactorily handle at least the common combination of scripts of non-Latin script and Latin, Digit, and Symbols script.

|       | SYM  | DIG  | LAT  | ARA | BENG | HANG | CJK  | HIR  | KAT |
|-------|------|------|------|-----|------|------|------|------|-----|
| SYM   | 4361 | 2285 | 3264 | 400 | 482  | 652  | 903  | 378  | 312 |
| DIG   | 2285 | 2838 | 2166 | 205 | 136  | 460  | 758  | 274  | 219 |
| LAT   | 3264 | 2166 | 5047 | 501 | 150  | 443  | 876  | 299  | 258 |
| ARA   | 400  | 205  | 501  | 797 | 0    | 0    | 0    | 0    | 0   |
| BENG  | 482  | 136  | 150  | 0   | 795  | 0    | 0    | 0    | 0   |
| HANG  | 652  | 460  | 443  | 0   | 0    | 847  | 81   | 32   | 28  |
| CJK   | 903  | 758  | 876  | 0   | 0    | 81   | 1615 | 447  | 355 |
| HIR   | 378  | 274  | 299  | 0   | 0    | 32   | 447  | 462  | 300 |
| KAT   | 312  | 219  | 258  | 0   | 0    | 28   | 355  | 300  | 374 |

Table 3: Script co-occurrence at image level on the ICDAR MLT-RRC 2017 [33] validation dataset. The row-column entry is incremented for all pairs of scripts present in an image.

|       | SYM  | DIG | LAT   | ARA  | BENG | HANG | CJK  | HIR  | KAT  |
|-------|------|-----|-------|------|------|------|------|------|------|
| LAT   | 1046 | 635 | 52050 | 0    | 0    | 0    | 36   | 0    | 0    |
| ARA   | 20   | 13  | 0     | 4881 | 0    | 0    | 0    | 0    | 0    |
| BENG  | 63   | 5   | 0     | 0    | 3688 | 0    | 0    | 0    | 0    |
| HANG  | 118  | 84  | 0     | 0    | 0    | 3767 | 0    | 0    | 0    |
| CKH   | 416  | 499 | 21    | 0    | 0    | 0    | 7265 | 1424 | 1037 |

Table 4: Script co-occurrence in words in the ICDAR MLT-RRC 2017 validation dataset [33]. Column: script of a character. Row: script / script group of the word the character appeared in. If multiple scripts are present in the word, the row-column entry is incremented for each script.

**The OCR accuracy on cropped words** of the E2E-MLT is shown in Tab. 5 and the confusion matrix for individual script in Tab. 6. In this evaluation, the ground truth for a word is defined as the most frequent script. Tab. 6 shows that E2E-MLT does not make many mistakes due to confusing script confusion.

| Script | Acc | $\frac{Edits}{len(GT)}$ | Character Instances | Images |
|--------|-----|------------------------|---------------------|--------|
| Symbol | 0.416 | 0.472 | 926 | 541 |
| Digit | 0.705 | 0.146 | 7027 | 1864 |
| Latin | 0.744 | 0.111 | 52708 | 9280 |
| Arabic | 0.462 | 0.250 | 4892 | 951 |
| Bengali | 0.342 | 0.314 | 3781 | 673 |
| Hangul | 0.652 | 0.217 | 3853 | 1164 |
| CJK | 0.446 | 0.295 | 7771 | 1375 |
| Hiragana | 0.317 | 0.268 | 1678 | 230 |
| Katakana | 0.130 | 0.434 | 1011 | 177 |
| Total | 0.651 | 0.164 | 83647 | 16255 |

Table 5: E2E-MLT OCR accuracy on the ICDAR MLT-RRC 2017 validation dataset [33]

| | Sym | Dig | Lat | Ara | Beng | Hang | CJK | Hir | Kat |
|------|-----|------|------|-----|------|------|------|-----|-----|
| Sym | 338 | 85 | 90 | 5 | 2 | 1 | 8 | 1 | 0 |
| Dig | 57 | 1695 | 87 | 9 | 4 | 1 | 6 | 1 | 2 |
| Lat | 172 | 92 | 8946 | 27 | 2 | 4 | 36 | 3 | 3 |
| Ara | 28 | 13 | 61 | 843 | 2 | 1 | 2 | 1 | 0 |
| Beng | 17 | 9 | 19 | 4 | 615 | 4 | 5 | 0 | 0 |
| Hang | 47 | 27 | 37 | 1 | 6 | 1013 | 34 | 2 | 1 |
| CJK | 47 | 13 | 14 | 0 | 0 | 3 | 1281 | 8 | 10 |
| Hir | 19 | 5 | 12 | 2 | 0 | 3 | 25 | 159 | 5 |
| Kat | 12 | 2 | 7 | 0 | 0 | 1 | 18 | 6 | 131 |

Table 6: Confusion matrix of E2E-MLT OCR on the ICDAR MLT-RRC 2017 [33] validation dataset. GT script is in row, the recognized script in columns

In Tab. 7 we show some difficult cases for multi-language text recognition and demonstrate that some of the mistakes are non-trivial for a human reader as well.

**Mistakes in the ground truth** annotations adds to the challenge, see Tab. 8. For Latin-script native readers, they are hard to identify. Another common source of error is caused by GT bounding boxes that are incorrectly axis-aligned bounding.

## 4.2 End-to-End Scene Text on the ICDAR 2015 dataset

The ICDAR 2015 dataset was introduced in the ICDAR 2015 Robust Reading Competition [21]. This dataset consists of 500 images taken from Google Glass device of people walking in city. Note that this dataset was not made with text in mind and thus adequately captures the real world scene text scenario. The dataset is referred as Incidental text and the text instances are small compared to entire image size. Unlike ICDAR RRC-MLT [33], the images in this dataset contain only English text.

Table 7: Difficult cases (for Latin-script native readers). Transcription errors, shown in red, which require close inspection - (a), (g), (h), (i). Note that for (i), the error is also in the ground truth. We were not able to establish a clear GT for (e) and (f). For (b), the transcription is 70004 in Bangla. In the context of Latin scripts, this same image will be interpreted as 900 08. Note the errors related to ":" and —, there are multiple types of colons and dashes in UTF.

Table 8: Errors in the GT of the ICDAR MLT-RRC 2017 validation dataset [33]. Incorrect transcriptions are highlighted in red. Note that some errors lead to very large edit distances, e.g. in (f) and (h). GT errors effect both training and evaluation of the method. We estimate that at least 10% of errors on non-latin words reported for E2E-MLT on the ICDAR MLT dataset are due to GT mistakes.

Here we experimentally evaluate E2E-MLT on ICDAR 2015 [21] dataset. Note that E2E-MLT is trained to handle multi-language text and the same is capable of handling text instances from other languages (we do not train a different model for English).

| Method | Recall | Precision | **F1-score** | Recall ED1 |
|---|---|---|---|---|
| TextProposals + DictNet* [11,16] | 33.61% | 79.14% | 47.18% | - |
| DeepTextSpotter* [5] | - | - | 47.0% | - |
| FOTS-Real time* [30] | - | - | 51.40% | - |
| FOTS-multi-scale* [30] | 53.20% | 84.61% | 65.33% | - |
| E2E-MLT | 44.9% | 71.4% | 55.1% | 59.6% |

Table 9: End-to-end evaluation on generic setting (no lexicon) of ICDAR 2015 [21] dataset. The methods marked with asterisk are trained only for English scene text. E2E-MLT has a single unconstrained (no dictionary) model for six languages (7,500 characters). Further, the testing of E2E-MLT is done using a single-scale of input images. The low precision of E2E-MLT is partially because the ICDAR 2015 dataset contains non-latin text as well.

E2E-MLT is a unconstrained multi-language scene text method. We do not make use of any per-image lexicon or dictionary. Thus, we evaluate E2E-MLT only for generic category where no per-image lexicon is available. Quantitative results are shown in Tab 9.

### 4.3   End-to-End Multi-Language Scene Text on the ICDAR MLT

We evaluate the performance of E2E-MLT on the validation set of ICDAR MLT 2017 [33] dataset. The dataset [33] comprises 7,200 training, 1,800 validation and 9,000 testing natural scene images. The ground truth annotations include bounding box coordinates, the script class and text-transcription. The dataset deals with the following 6 languages: Arabic, Latin, Chinese, Japanese, Korean, Bangla. Additionally, punctuation and some math symbols sometimes appear as separate words and are assigned a special script class called Symbols, hence 7 script classes are considered. Quantitative evaluation of end-to-end recognition (localization and recognition) is shown in Tab. 10. Qualitative results are demonstrated in Tab. 11.

| Text Length | E2E Recall | Precision | E2E Recall ED1 | Loc. Recall IoU 0.5 |
|---|---|---|---|---|
| 2+ | 42.9% | 53.7% | 55.5% | 68.4% |
| 3+ | 43.3% | 59.7% | 59.9% | 69.5% |

Table 10: E2E-MLT end-To-end recognition results on ICDAR MLT 2017 [33] validation set.

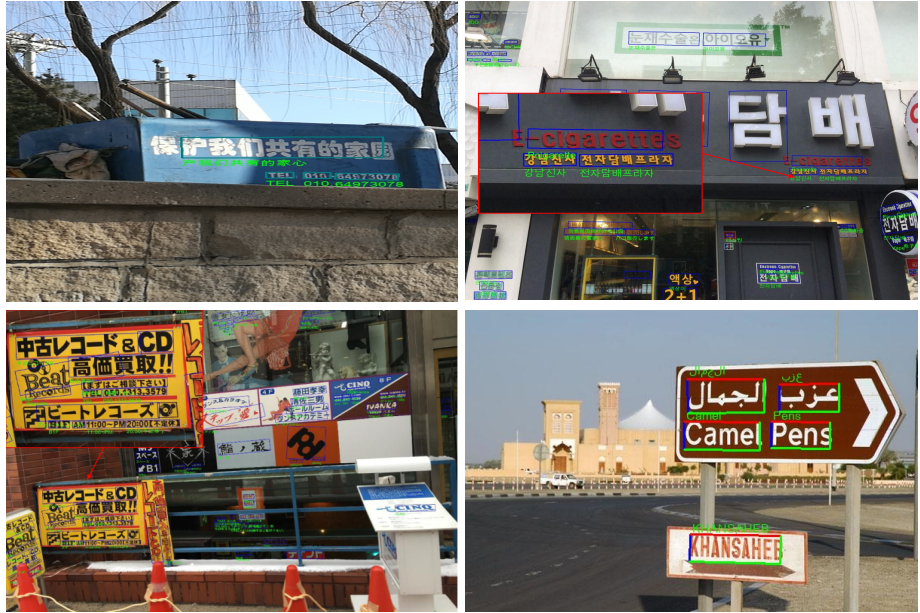

Table 11: Example E2E-MLT results on the ICDAR MLT 2017 dataset [33]

### 4.4  Joint Multi-Language Text Localization and Script Identification on the ICDAR RRC-MLT 2017

The objective of this task is to predict precise word level bounding boxes and the corresponding script class for each word. Existing text recognition algorithms [18, 19, 39] are language-dependent which makes script identification a prerequisite task for the other methods. In Section 4.3, we experimentally demonstrated that script identification is not required for multi-language text recognition. In E2E-MLT, once we transcribe the text, we make use of a simple majority voting from each character to predict the final script class.

In Tab. 12, we provide quantitative evaluation on joint text localization and script identification task of ICDAR MLT 2017 [33]. Note that unlike other methods in Tab. 12, we do not solve script identification as a separate task. Rather, we make use of transcribed text output of E2E-MLT.

| Method | F-Measure | Recall | Precision |
|---|---|---|---|
| SCUT-DLVClab2 | 58.08% | 48.77% | 71.78% |
| TH-DL | 39.37% | 29.65% | 58.58% |
| E2E-MLT | 48.00% | 45.98% | 50.20% |

Table 12: Joint text localization and script identification on the ICDAR RRC-MLT [33] test data.

## 5   Conclusion

E2E-MLT, an end-to-end trainable unconstrained method for multi-language scene text localization and recognition, was introduced. It is the first published multi-language OCR for scene text capable of recognizing highly rotated and vertical text instances.

Experimentally, we demonstrate that a single FCN based model is capable of jointly solving multi-language scene text localization and recognition tasks. Although E2E-MLT is trained to work across six languages, it demonstrates competitive performance on ICDAR 2015 [21] dataset compared to the methods trained on English data alone.

With the paper we will publish a large scale synthetically generated dataset for training multi-language scene text detection, recognition and script identification methods. The code and trained model will be released.

## 6   Acknowledgement

The research was supported by the Czech Science Foundation Project GACR P103/12/G084.